\documentclass[10pt,twocolumn,letterpaper]{article}

\usepackage{cvpr}              
\usepackage{booktabs}
\usepackage{pifont}
\usepackage{graphicx}
\usepackage{subcaption}
\usepackage{multirow}
\usepackage{makecell, multicol, diagbox}
\usepackage{xcolor, colortbl}
\usepackage{array}
\usepackage[accsupp]{axessibility}  
\newcommand{\revise}[1]{{\color{blue}#1}}

\newcommand{\method}{RNG\xspace}
\newcommand{\rgs}{GS$^3$\xspace}

\usepackage{pifont}
\usepackage{graphicx}
\usepackage{subcaption}
\usepackage{multirow}
\usepackage{makecell, multicol, diagbox}
\usepackage{xcolor, colortbl}
\usepackage{array}
\newcolumntype{C}[1]{>{\centering\arraybackslash}p{#1}}
\newcommand{\winner}[1]{\cellcolor{red!20}{#1}}
\newcommand{\runner}[1]{\cellcolor{orange!20}{#1}}

\newcommand{\cellwidth}{0.07}

\def\figurePath{fig/}
\def\myfigure#1#2#3{
\begin{figure}[tb]\centering\includegraphics[width = \linewidth]{\figurePath#2}\caption{#3}\label{fig:#1}
\end{figure}}
\def\mycfigure#1#2#3{
\begin{figure*}[htb]\centering\includegraphics*[clip, width = \linewidth]{\figurePath#2}\caption{#3}\label{fig:#1}
\end{figure*}}

\usepackage{gensymb}

\definecolor{cvprblue}{rgb}{0.21,0.49,0.74}
\usepackage[pagebackref,breaklinks,colorlinks,allcolors=cvprblue]{hyperref}

\def\paperID{5355} 
\def\confName{CVPR}
\def\confYear{2025}

\title{\method: Relightable Neural Gaussians }

\author{
Jiahui Fan$^1$,
\ 
Fujun Luan$^2$,
\ 
Jian Yang$^1$\thanks{Corresponding authors.}
\ 
Milo\v{s} Ha\v{s}an$^2$, and Beibei Wang$^{3*}$
\\
$^1$PCA Lab
\thanks{
PCA Lab, Key Lab of Intelligent Perception and Systems for High-Dimensional Information, School of Computer Science and Engineering, Nanjing University of Science and Technology, China. 
}, 
Nanjing University of Science and Technology\\
$^2$Adobe Research\ 
$^3$Nanjing University
\\
{\tt\small \{fjh, csjyang\}@njust.edu.cn}, 
{\tt\small fluan@adobe.com}, 
{\tt\small milos.hasan@gmail.com}, 
{\tt\small beibei.wang@nju.edu.cn}
}

\begin{document}
\maketitle

\begin{abstract}
3D Gaussian Splatting (3DGS) has shown impressive results for the novel view synthesis task, where lighting is assumed to be fixed. However, creating relightable 3D assets, especially for objects with ill-defined shapes (fur, fabric, etc.), remains a challenging task. The decomposition between light, geometry, and material is ambiguous, especially if either smooth surface assumptions or surface-based analytical shading models do not apply. We propose Relightable Neural Gaussians (\method), a novel 3DGS-based framework that enables the relighting of objects with both hard surfaces or soft boundaries, while avoiding assumptions on the shading model. We condition the radiance at each point on both view and light directions. We also introduce a shadow cue, as well as a depth refinement network to improve shadow accuracy. Finally, we propose a hybrid forward-deferred fitting strategy to balance geometry and appearance quality. Our method achieves significantly faster training (1.3 hours) and rendering (60 frames per second) compared to a prior method based on neural radiance fields and produces higher-quality shadows than a concurrent 3DGS-based method. Project page: \url{whois-jiahui.fun/project_pages/RNG}.
\end{abstract}    
\section{Introduction}

Creating 3D assets from multi-view captures of the real world is an effective way for content creation, avoiding manual modeling labor. The resulting 3D assets can be objects with well-defined surfaces or ill-defined shapes (e.g., fur, fabric, grass, etc.), as both are important in many applications. Unfortunately, if we want the resulting assets to be \emph{relightable}, the task is still challenging because of the ill-posed nature of the decomposition between light, materials, and geometry. This is especially true for complex non-smooth materials, which raise difficulties in decomposition, as surface-specific constraints or surface-based analytical shading models cannot be leveraged. In this paper, we aim to relight objects with clear surfaces or soft boundaries given multi-view captured images with varying illumination, simultaneously achieving high-quality relighting and shortening training/rendering times. 

After ground-breaking view-synthesis work on Neural Radiance Fields (NeRF)~\cite{mildenhall_2020_nerf} and 3D Gaussian Splatting (3DGS)~\cite{kerbl_2023_3dgs}, extensive efforts have focused on reconstructing relightable 3D assets \cite{wang_2021_neus, liu_2023_nero, li_2024_tensosdf, jin_2023_tensoir, gao_2023_relightablegs, jiang_2024_gaussianshader, liang_2024_gsir}. However, these methods mostly rely on surface shading models and introduce surface constraints (including the assumption of valid normals), preventing them from reconstructing  objects with soft boundaries and/or materials that are not well represented with simple analytic models. Recently, NRHints~\cite{zeng2023relighting} enabled relightable capture of both smooth surfaces and objects with soft boundaries by using input views with a moving point light and a neural appearance model. Being based on a NeRF framework, NRHints suffers from high training/rendering time costs and some over-smoothing of detail. The concurrent work \rgs~\cite{bi2024rgs} uses the same capture setup but instead uses 3DGS as the underlying framework, which is more efficient in training and rendering, and captures finer details. With the less accurate geometry obtained from 3DGS, \rgs has relatively lower shadow quality. 

In this paper, we propose \emph{Relightable Neural Gaussians} (\method), a novel 3DGS-based framework for relighting objects with both clear surfaces and soft boundaries. We implicitly model the radiance of objects by learning latent (feature) vectors at each \textit{neural Gaussian}. To interpret neural Gaussians, we use the \textit{neural Gaussian decoder} network, and condition it on the view \textit{and light} directions. Analytical assumptions in shading models and surface constraints are avoided in our neural representation, making it capable of learning appearances that do not fit well into those constraints.

Following prior work~\cite{zeng2023relighting}, we utilize views with a moving point light, to observe many view/light combinations and reduce ambiguities in decomposition. However, point lights produce sharp shadows, which are challenging for neural networks to capture accurately. We present a \emph{shadow cue with depth refinement} to condition the neural Gaussian decoder, improving the shadow quality. We also introduce a two-stage hybrid (forward-deferred) optimization pipeline for better shadow appearance. 

In our results, \method shows not only higher-quality details than the NeRF-based prior method NRHints, but also more accurate shadows than the concurrent 3DGS-based approach \rgs. In terms of performance, \method takes about 1.3 hours for training and achieves a $60$ frame per second (fps) rendering performance on an RTX 4090 GPU, which is competitive with 3DGS and \rgs and many times faster than NRHints.

To summarize, our main contributions include 
\begin{itemize}
\item a relightable neural Gaussian framework to render objects with smooth surfaces or soft boundaries, under arbitrary view and light directions, and with no analytic assumptions on the shading model,

\item a shadow cue technique and a depth refinement network to enhance the quality of shadows, and

\item a hybrid (forward-deferred) optimization strategy, achieving high-quality reconstruction and sharp shadow appearance.
\end{itemize}

\section{Related work}

\subsection{Inverse rendering}

Inverse rendering \cite{bi_2020_neural, zhang_2021_physg, liu_2023_nero} decomposes the light, material, and geometry with multi-view RGB inputs, and the decomposed assets can be relit under any desired novel lighting. To represent the materials, several methods introduce a standard shading model similar to the Disney Principled BRDF~\cite{burley_2012_physically} as a physically-based prior, and neural materials~\cite{zhang_2021_nerfactor} can also be utilized for material recovery.  
With the representation capacity of NeRF, some methods~\cite{, boss_2021_neuralpil, boss_2021_nerd, yang_2023_sireir, zhang_2023_neilfpp} produce high-quality inverse rendering at the cost of high training consumption and slow rendering speed. Jin et al.~\cite{jin_2023_tensoir} use grid features to represent the scenes, leading to relatively fast training speed. Zhang et al.~\cite{zhang2023nemf} adopts the SGGX Microflake model~\cite{heitz2015sggx} to perform inverse rendering, achieving unique effects for semi-transparent targets. 
SDFs are also commonly used in the representations~\cite{li_2024_tensosdf, wang_2021_neus, zhang_iron_2022}, leading to smoother surface normals but biasing the method to objects with relatively smooth surfaces.

3DGS brings the rasterization framework into multi-view stereo reconstructions. However, this nature of 3DGS also hurts the quality of its obtained geometric attributes, such as depth and normal, making them noisy and difficult for further use. Some existing methods~\cite{huang_2024_2dgs, dai_2024_gaussiansurfels,guedon_2024_sugar, chen_2023_neusg} bring constraints or introduce meshes into the Gaussians, improving the geometry quality. 
By introducing analytical shading models, several works utilize 3DGS to achieve inverse rendering under unknown environment lighting. Jiang et al.~\cite{jiang_2024_gaussianshader} and Shi et al.~\cite{shi_2023_gir} supervise the normals via the orientation of Gaussians, while Gao et al.~\cite{gao_2023_relightablegs} and Liang et al.~\cite{liang_2024_gsir} leverage depth to obtain the normal information.

Compared to NeRF-based methods, 3DGS-based methods are more efficient and handle soft-boundary objects better due to their flexible representation. Therefore,  we choose to use 3DGS as the underlying framework. Further, previous approaches with surface priors fail to handle soft objects, and the fixed types of analytical models also limit the application. In contrast, our method generalizes across a wider range of scenarios without such assumptions.

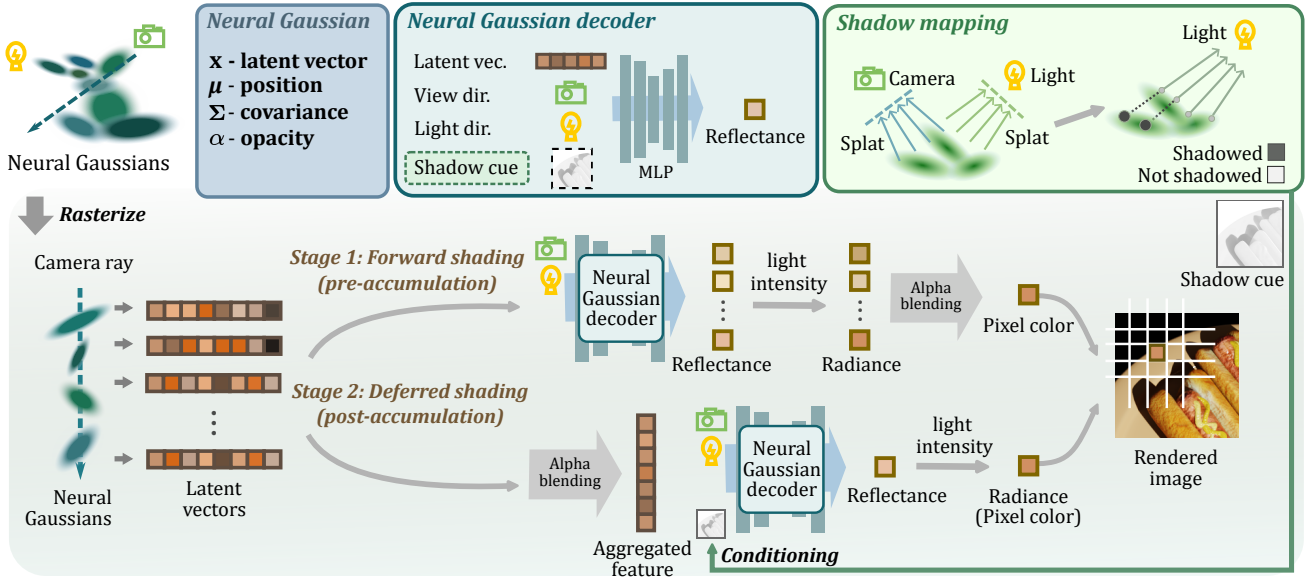
\mycfigure{pipeline}{pipeline_2.pdf}{The overview of \method. Each Gaussian point in the scene contains an extra latent vector that describes the reflectance. The latent values interpreted by an MLP decoder, conditioned on view and light directions.  Training has two stages. In the first stage, we employ \textit{forward shading}, where we decode all the latent vectors of Gaussian points into colors, followed by the alpha blending. In the second \textit{deferred shading} stage, we first alpha-blend the neural Gaussian features to get an aggregated feature, and then we feed it to the decoder. We apply shadow mapping to obtain a shadow cue map and use the shadow cue as an extra input for the decoder in the second stage. }

\subsection{Relighting of ill-defined shapes}

Most existing inverse rendering methods cannot simply extend to soft-boundary objects, due to the incompatibility of shading models and the challenging light transport in such scenes.  Gao et al.~\cite{gao2020deferred} present Deferred Neural Relighting, which leverages learned neural texture on a rough proxy geometry for relighting objects including fluffy shapes. 
Mullia et al.~\cite{mullia2024rna} propose a novel representation that combines explicit geometry with a neural feature grid and an MLP decoder, achieving high-fidelity rendering and relighting with good flexibility and integration. Unlike our method, they only support synthetic inputs and require ground-truth geometry. 
\revise{
}

Recently, Zeng et al.~\cite{zeng2023relighting} proposed NRHints, which maintains an implicit neural representation with both SDF and NeRF-style feature grids, and predicts radiance with shadow and highlight hints, achieving high-quality relighting. However, NRHints is computationally heavy in training and rendering, and tends to over-smooth soft objects, especially at boundaries. 
Our concurrent work \rgs~\cite{bi2024rgs} introduces triple splatting to relight objects. They introduce analytical appearance approximation that is also supplemented by neural networks, enabling high-efficiency relighting for fluffy objects as well. \rgs still suffers from the inherent lower-quality geometry of Gaussian point cloud, leading to less sharp shadow appearance.

We target the same problem as in NRHints and \rgs, and use point-lit images as input as well. However, we introduce neural Gaussians, avoiding surface constraints and shading model assumptions, gaining more flexibility in representation. We further propose the shadow cue with depth refinement to enhance the shadow quality, and design a hybrid optimization strategy. Overall, \method achieves faster training/rendering and finer details than NRHints, as well as higher shadow quality under point lights than \rgs.


\section{Method }

The goal of our work is to reconstruct high-quality relightable assets for objects with both hard surfaces and soft boundaries while maintaining fast training and rendering time. We propose relightable neural Gaussians  (\cref{sec:neural}) to implicitly model the reflectance. We also apply a shadow cue with depth refinement (\cref{sec:shadow}) to improve the quality of shadows and design a hybrid forward-deferred optimization strategy (\cref{sec:pipeline}) to further improve the shadow appearance while preserving the quality of geometry. Fig.~\ref{fig:pipeline} illustrates the overview of our method.

%
\subsection{Background: 3D Gaussian Splatting}
\label{sec:background}

3DGS represents a scene with a set of 3D Gaussians, each of which is defined as
\begin{equation}
    \mathrm{Gaussian}(x|\mu,\Sigma) = e^{-\frac{1}{2}(x-\mu)^T\Sigma^{-1}(x-\mu)},
\end{equation}
where $x$ is a position in the scene, $\mu$ is the mean of the Gaussian, and $\Sigma$ denotes the covariance matrix of the 3D Gaussian, which is factorized into a scaling matrix $S$ and a rotation matrix $R$ as $\Sigma=RSS^TR^T$. To render an image, 3DGS projects the 3D Gaussians onto the 2D image plane and employs alpha blending on the sorted Gaussians as 
\begin{equation}
    C = \sum_{i\in \mathbb{N}}{c_i\alpha_i\prod_{j=1}^{i-1}(1-\alpha_j)},
    \label{eq:gsblend}
\end{equation}
where $c_i$ is the color of each Gaussian, and $\alpha_i$ is given by evaluating a projected 2D Gaussian with covariance $\Sigma^\prime$ multiplied with a learned per-point opacity. In 3DGS, the alpha blending of color $c_i$ (which depends on the view direction and is represented by spherical harmonics) from every nearby Gaussian point yields the reflectance at position $x$.

\subsection{Relightable neural Gaussians}
\label{sec:neural}

Existing 3DGS-based relighting methods leverage analytical shading models and/or surface assumptions. Instead, we use a learned latent space to implicitly represent the view- and light-dependent reflectance in the scene. As shown in Fig~\ref{fig:pipeline}, each Gaussian point carries a latent (feature) vector that models this reflectance. 
To enable relightability, the reflectance has to be dependent on not only the view directions $\mathbf{\omega_{o}}$ but also the light directions $\mathbf{\omega_{i}}$. Therefore, the network can decode and predict the reflectance values at novel light positions. The final reflectance is represented as
\begin{equation}
\label{eq:neural}
    \rho(\mathbf{x}, \mathbf{\omega_o}, \mathbf{\omega_i}) = \Theta(\mathbf{x}|\mathbf{\omega_o}, \mathbf{\omega_i}),
\end{equation}
where $\Theta$ is the neural Gaussian decoder and $\mathbf{x}$ is the shading point with its corresponding latent vector. This reflectance value is analogous to the BRDF times cosine term from the standard rendering equation, and needs to be multiplied by light intensity and light falloff to obtain final radiance from a point light. 
When applying novel lighting conditions, the network takes the given point light positions and view directions as conditioning inputs, leading  (after combining with incoming light intensity) to a neural implicit relightable radiance representation.

%
\subsection{Shadow cue}
\label{sec:shadow}

With our proposed neural Gaussians, the reflectance at positions in the scene can be represented by latent vectors stored in each Gaussian point. However, there are still some potential quality issues. First, the network tends to over-fit all view/light directions in the training set, resulting in blurry or incomplete shadows in unseen predictions or inconsistent shapes in movement. Second, point lights yield sharp shadows, and the MLP is prone to over-smooth such high-frequency signals. 

To address the above issues, we introduce a shadow cue to condition the neural Gaussian decoder. The shadow cue is a 1-channel map in the screen space that indicates the visibility to the light of each shading point and will be fed into the MLP together with other inputs described in Eq.~\ref{eq:neural}. 

We obtain the shadow cue by performing shadow mapping under the 3DGS framework. Shadow mapping requires the precise locations of shading points. However, since we do not explicitly trace rays, we can instead use the depth value for each pixel for shading point computation. Obtaining the depth values of a Gaussian cloud is not well-defined. Therefore, we introduce a depth refinement network to correct the depth values and help find the valid shading points.

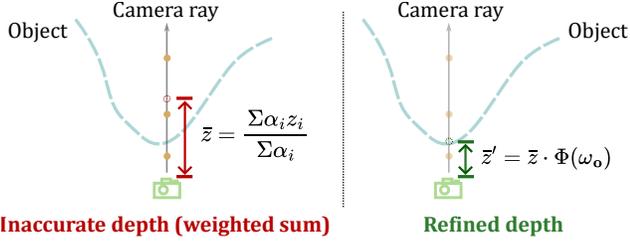
\myfigure{depth_refine}{depth_refine.pdf}{The effect of the depth refinement network. The weighted sum of Gaussian depths is not accurate, resulting in mismatching shadow cues. Therefore, we propose a depth refinement network to correct the depth.}

\paragraph{Depth refinement.}
An intuitive and naive proxy for the depth is the weighted sum of the depth and Gaussian responses along the ray,
\begin{equation}
    \bar{z} = \frac{\Sigma \alpha_i z_i}{\Sigma \alpha_i},
\end{equation}
where $z_i$ is the depth value of $i^\textrm{th}$ Gaussian on the camera ray and $\alpha_i$ is the ray response at each intersections of 3D Gaussians. Usually, this proxy is normalized, thus the background leakage at semi-transparent areas can be reduced. However, sometimes the weighted sum is incorrect, leading to wrongly located shading points and consequently mismatching shadow cues. We discuss and showcase this situation in the supplementary. To address this issue, we propose a depth refinement network to correct the shading point locations by learning a scaling factor, as shown in Fig.~\ref{fig:depth_refine}.  We assume the depth correction is linear for each pixel and dependent on view directions $\omega_o $. Therefore, the refined depth value is obtained by 
\begin{equation}
    \bar{z}^\prime = \bar{z} \cdot \Phi( \omega_o ),
\end{equation}
where $\Phi$ is the depth refinement MLP.

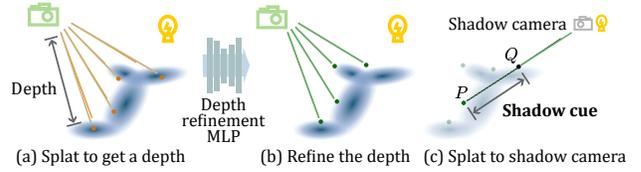
\myfigure{shadow}{shadow.pdf}{The illustration of shadow cue computation. First, we splat the Gaussians onto the camera to get depth values. Then, we run the depth refinement network to correct them and locate the shading points $P$. At last, we splat the shading points onto the shadow camera to find the intersections of shadow rays $Q$, and store the distance $\vert PQ \vert$ as the shadow cue.}

\paragraph{Shadow mapping.}
Conventional shadow mapping marches the ray to get visibility, which can be expensive and difficult to achieve. Instead, under the 3DGS framework, we perform an extra pass of Gaussian splatting to cast shadow mapping. We obtain the shadow cue in the following steps, as shown in Fig.~\ref{fig:shadow}. First, we splat Gaussians onto the camera and record the depth values of each pixel. Second, we run the depth refinement network to correct the depth and calculate a shading point $P$ for each pixel based on the pixel depth. After this, we set a virtual shadow camera at the point light position. We splat Gaussians onto the shadow camera for a second pass, recording the depth to find a shading point $Q$ for each pixel in the shadow camera. We project $P$ into the shadow camera frame to find its corresponding $Q$.
Since $Q$ is equivalently the shadow ray intersection, the distance $\vert{PQ}\vert$ is recorded as the shadow cue for this pixel. 

Note that for computational efficiency, we omit the depth refinement when we obtain the shadow camera depth. With shadow cues, the neural Gaussian decoder takes multiple inputs, and all of them contribute to the final color of a single Gaussian point. The final reflectance is represented as 
\begin{equation}
     \rho(\mathbf{x}, \mathbf{\omega_o}, \mathbf{\omega_i}) = \Theta(\mathbf{x}|\mathbf{\omega_o}, \mathbf{\omega_i}, V),
\end{equation}
where $V=\vert PQ \vert$ is the shadow cue. In practice, we use the same resolution as the camera for the shadow camera, and we apply a clamping between zero and the scene units to the shadow cue map for the stability of training.

\subsection{Hybrid optimization}
\label{sec:pipeline}

With all the components above, we now have the \method framework, where the scene is represented as a structure of neural Gaussian points, and the reflectance at each Gaussian point is represented as a feature vector that is conditioned on view/light directions and shadow cues. The adjusted 3DGS rasterization operation becomes
\begin{equation}
    C_{\mathrm{forward}} = \sum_{i\in \mathbb{N}}{\Theta(\mathbf{x_i}|\mathbf{\omega_o}, \mathbf{\omega_i}, S, V)\alpha_i\prod_{j=1}^{i-1}(1-\alpha_j)},
    \label{eq:gsblend}
\end{equation}
where $C_\mathrm{forward}$ is the color at each pixel. We call this rasterization procedure \textit{forward shading}.

In forward shading, the alpha blending after the reflectance computation blurs the shadow. To address this problem, we introduce the \textit{deferred shading}.

\paragraph{Deferred shading.}
In deferred shading, we blend the feature vectors of Gaussians first to get an aggregated feature in image space, and then we decode it with the neural Gaussian decoder. In this case, we propose the rasterization of deferred shading as

\begin{equation}
    C_\textrm{defer} = \Theta(\sum_{i\in \mathbb{N}}{\mathbf{x_i}\alpha_i\prod_{j=1}^{i-1}(1-\alpha_j)} | \mathbf{\omega_o}, \mathbf{\omega_i}, V).
\end{equation}

To our observation, forward shading produces better geometry and worse shadow, while deferred shading improves the shadow appearance but causes floaters.  Therefore, we design a two-stage hybrid optimization strategy to benefit from both options. Further investigation into this choice is discussed in the supplementary.

\paragraph{Two-stage strategy.}
The whole training procedure of \method consists of two stages. We employ forward shading in the first stage to get Gaussian points and latent vectors and use deferred shading in the second stage. In the second stage, we enable the shadow cue and re-train the neural Gaussian decoder. We keep all learned latent vectors in the first stage as initialization of the second stage, in order to provide more semantic information and accelerate the training of the second stage.
Note that in the first stage, we do not enable the shadow cue, because at the early stage, the Gaussians are not well-shaped, and the generated wrong shadow information may hurt the training stability. 

\begin{table*}[htb]
    \centering
    \caption{Comparison of various scenes between NRHints~\cite{zeng2023relighting}, \rgs~\cite{bi2024rgs} and our method. We provide (from left to right) PSNR~$(\uparrow)$, SSIM~$(\uparrow)$, and LPIPS~$(\downarrow)$ for comparison, and the best/second-best results are colored in \colorbox{red!20}{red}/\colorbox{orange!10}{orange}, respectively. 
    We produce the best or second-best results on most scenes and with better PSNR, SSIM, and LPIPS values on average, indicating high-fidelity reconstruction and realistic details in our renderings. 
    }
    \label{tab:metrics}
    \begin{tabular}{p{0.10\textwidth} C{\cellwidth\textwidth} C{\cellwidth\textwidth} C{\cellwidth\textwidth} C{\cellwidth\textwidth} C{\cellwidth\textwidth} C{\cellwidth\textwidth} C{\cellwidth\textwidth} C{\cellwidth\textwidth} C{\cellwidth\textwidth}}
        \toprule
        Scene & \multicolumn{3}{c}{NRHints\cite{zeng2023relighting}} & \multicolumn{3}{c}{\rgs\cite{bi2024rgs}} & \multicolumn{3}{c}{Ours} \\
        \midrule
        
        Cat        & 
        \runner{28.3712} & 0.8751 & 0.1318 &
        26.0850 & \runner{0.8815} & \runner{0.1019} &
        \winner{28.3869} & \winner{0.8883} & \winner{0.0847} \\ 
        
        CatSmall   & 
        \winner{35.4472} & \runner{0.9705} & 0.0450 &
        34.4018 & \winner{0.9729} & \runner{0.0390} & 
        \runner{34.7511} & 0.9699 & \winner{0.0377} \\ 
        
        Cluttered   & 
        \winner{32.1470} & {0.9434} & 0.0629 &
        {30.2874} & \winner{0.9443} & \runner{0.0489} &
        \runner{30.7970} & \runner{0.9442} & \winner{0.0456} \\
        
        CupFabric  & 
        \runner{38.1833} & \runner{0.9831} & 0.0256 &
        37.1364 & 0.9830 & \runner{0.0236} &
        \winner{38.5429} & \winner{0.9857} & \winner{0.0170} \\
        
        Fish       & 
        30.2113 & 0.9000 & 0.1176 &
        \runner{30.8571} & \runner{0.9180} & \runner{0.0668} & 
        \winner{31.0113} & \winner{0.9195} & \winner{0.0561} \\
        
        FurBall    & 
        \runner{26.7098} & \winner{0.9340} & \runner{0.0524} & 
        26.3552 & \runner{0.9309} & 0.0577 & 
        \winner{27.8211} & 0.9263 & \winner{0.0436} \\ 
        
        HairBlonde & 
        32.4589 & 0.9497 & 0.0388 &
        \runner{32.9148} & \runner{0.9715} & \runner{0.0194} &
        \winner{34.7907} & \winner{0.9731} & \winner{0.0147 } \\
        
        Hotdog     & 
        \winner{32.8954} & \winner{0.9728} & \winner{0.0227} & 
        25.4029 & 0.9489 & 0.0483 & 
        \runner{30.3820} & \runner{0.9603} & \runner{0.0339} \\ 
        
        Lego       & 
        \winner{29.5974} & \winner{0.9559} & \winner{0.0300} & 
        26.6257 & 0.9226 & 0.0514 &
        \runner{26.7235} & \runner{0.9244} & \runner{0.0506} \\
        
        Pikachu    & 
        \winner{33.5846} & \winner{0.9716} & \winner{0.0248 } &
        \runner{32.1464} & \runner{0.9697} & {0.0294} &
        31.3826 & 0.9661 & \runner{0.0289} \\
        
        Pixiu      & 
        \winner{31.4333} & 0.9360 & 0.0751 &
        30.3765 & \runner{0.9371} & \runner{0.0640} &
        \runner{30.3485} & \winner{0.9410} & \winner{0.0540} \\
        
        RedCloth   & 
        \runner{34.0962} & 0.9186 & 0.1002 &
        31.6039 & \runner{0.9328} & \runner{0.0489}  &
        \winner{35.2186} & \winner{0.9489} & \winner{0.0282 } \\
        
        WhiteFur   & 
        23.4099 & 0.8871 & 0.1004 &
        \runner{32.8326} & \runner{0.9662} & \runner{0.0220 } &
        \winner{33.7007} & \winner{0.9700} & \winner{0.0140 } \\
        
        \midrule
        
        \textit{Average} & 
        \runner{31.4266} & 0.9383 & 0.0636 &
        30.5404 & \runner{0.9446} & \runner{0.0478} &
        \winner{31.8352} & \winner{0.9475} & \winner{0.0392} \\

        \bottomrule
    \end{tabular}
\end{table*}

\renewcommand{\cellwidth}{0.06}
\begin{table*}[htb]
    \centering
    \caption{Comparison of relighting under novel environment lighting with prior GS-based relighting methods~\cite{liang_2024_gsir, gao_2023_relightablegs}. We provide (from left to right) PSNR~$(\uparrow)$, SSIM~$(\uparrow)$ and LPIPS~$(\downarrow)$ for comparison, and the best/second-best results are colored in \colorbox{red!20}{red}/\colorbox{orange!10}{orange}, respectively. 
    Our model significantly improves the accuracy in decomposing light and materials, yielding overall prevailing metrics.
    }
    \label{tab:relight_envmap}
    \begin{tabular}{p{0.10\textwidth} C{\cellwidth\textwidth} C{\cellwidth\textwidth} C{\cellwidth\textwidth} C{\cellwidth\textwidth} C{\cellwidth\textwidth} C{\cellwidth\textwidth} C{\cellwidth\textwidth} C{\cellwidth\textwidth} C{\cellwidth\textwidth}}
        \toprule
        Scene & \multicolumn{3}{c}{GS-IR\cite{liang_2024_gsir}} & \multicolumn{3}{c}{RelightableGS\cite{gao_2023_relightablegs}} & \multicolumn{3}{c}{Ours} \\
        \midrule
        
        Armadillo & 
        \runner{30.4157} & 0.8726 & \runner{0.0316} & 
        24.4747 & \runner{0.8765} & 0.0327 
        & \winner{37.0618} & \winner{0.9062} & \winner{0.0145} \\
        
        CupPlane & 
        20.7918 & 0.8577 & 0.0640 & 
        \runner{25.5422} & \runner{0.9083} & \runner{0.0292} & 
        \winner{28.7075} & \winner{0.9189} & \winner{0.0281} \\
        
        HairBlue & 
        \runner{26.6556} & 0.8154 & 0.0796 & 
        20.7416 & \runner{0.8187} & \runner{0.0721} & 
        \winner{31.3762} & \winner{0.8731} & \winner{0.0700} \\
        
        HairYellow & 
        23.2272 & 0.7996 & 0.1308 & 
        \runner{25.3962} & \runner{0.8266} & \runner{0.1152} & 
        \winner{25.6469} & \winner{0.8527} & \winner{0.1102} \\
        
        \midrule
        
        \textit{Average} & 
        \runner{25.2726} & 0.8363 & 0.0765 & 
        24.0387 & \runner{0.8575} & \runner{0.0623} & 
        \winner{30.6981} & \winner{0.8877} & \winner{0.0557} \\
        
        \bottomrule
    \end{tabular}
\end{table*}

\mycfigure{vs_nrhints_real+syn}{vs_nrhints_real+syn-SMALL.pdf}{Comparison between NRHints~\cite{zeng2023relighting}, \rgs~\cite{bi2024rgs} and our method on real/synthetic datasets under point lights. The best/second-best results are marked as \textbf{bold}/\textit{italic}, respectively. Our method has the lowest LPIPS with the shown images and is also the best or second-best in PSNR and SSIM values. Our method also has better shadow areas than \rgs.}

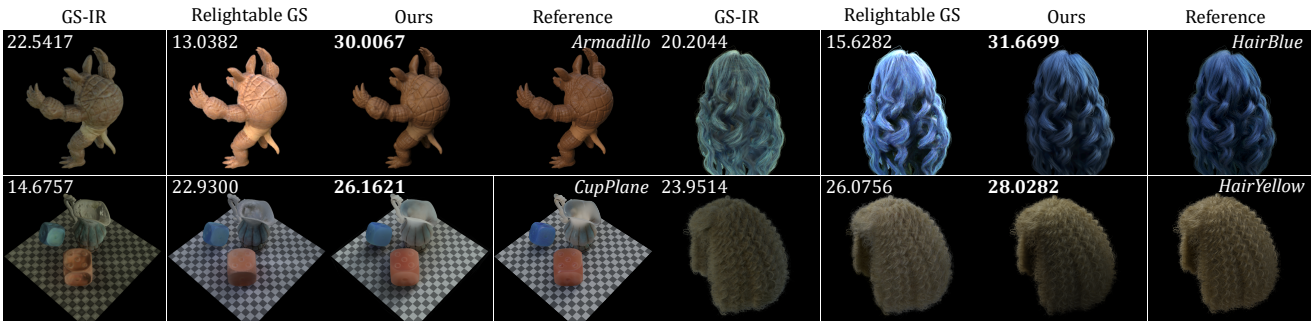
\mycfigure{relight_envmap}{relight_envmap-SMALL.pdf}{Comparison of relighting results under environment lighting with PSNR values of each image. We compare our relighting results with some previous GS-based methods \cite{liang_2024_gsir, gao_2023_relightablegs} and the ground truth. Our method decomposes the light and materials better and achieves better relighting, as we utilize point-lit images and the neural appearance model.}

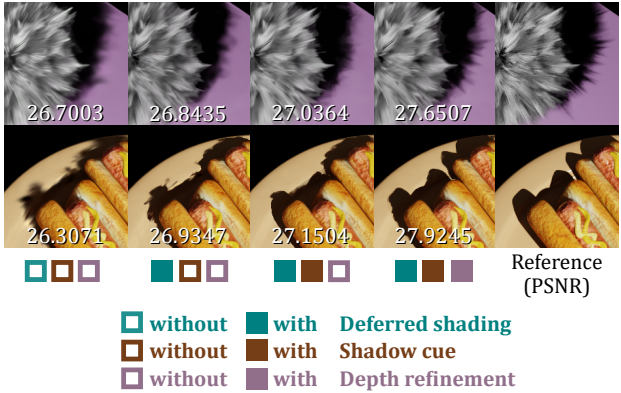
\myfigure{ablation-all}{ablation-all-SMALL.pdf}{The ablation study of \method components. We gradually remove them from our full model and show the quality gap between them with PSNR values. The shadow quality is significantly decreased without these components, and the PSNR values also demonstrate their effectiveness and necessity.}

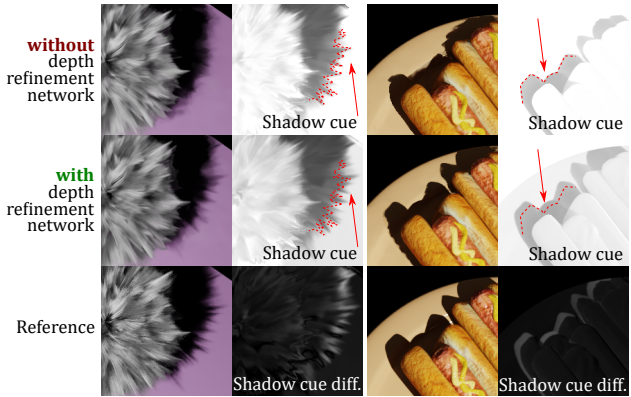
\myfigure{ablation-d_mlp}{ablation-d_mlp-SMALL.pdf}{The comparison of results with/without depth refinement MLP and the visualizations of corresponding shadow cues. The corresponding positions of shadow cues in both cases are marked with red arrows and dotted lines. With the depth refinement, the shadow mapping gives more reasonable and matched shadow cues, helping the network to better condition the appearance of the shadow information.}

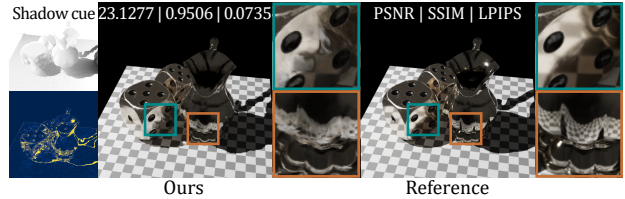
\myfigure{limitation}{limitation.pdf}{Comparison of our method and the ground truth on highly reflective objects. Our result blurs the reflection, as we did not introduce ray marching into our framework. }

\section{Results}

In this section, we validate the effectiveness and quality of \method. We first provide the implementation details in \cref{sec:implementation}, and describe about the experiment setups in \cref{sec:setup}. Then, we evaluate our method with quantitative results in \cref{sec:validation} and provide ablation studies in \cref{sec:albation}.

\subsection{Implementation}
\label{sec:implementation}

We implemented our method using the Pytorch~\cite{paszke2019pytorch} framework. The feature vector in each Gaussian is 16-channel, and the neural Gaussian decoder is an MLP with $4$ hidden layers and $256$ hidden units. We apply frequency encoding to both view/light directions and shadow cues, making them $15$ dimensions and $17$ dimensions, respectively. 
We use the Adam optimizer \cite{kingma_2014_adam} and train it at a learning rate $1.0 \times 10^{-3}$ for the color decoder MLP, $3.0 \times 10^{-4}$ for the depth refinement MLP and $2.5\times 10 ^ {-3}$ for feature vectors in Gaussian points. The same loss functions from 3DGS~\cite{kerbl_2023_3dgs} are used, which is a combination of $L_1$ loss and structural similarity index (SSIM)~\cite{wang_2004_ssim}. 
We train the model for a total $100$k steps, and the forward shading stage usually takes the first $30$k iterations. 
In order to improve the computational efficiency, we cache the shadow cue for each training view and only update them every $5$ iterations. 
We run all our results on an RTX 4090 GPU and i9-13900K CPU, powered by a Windows Subsystem Linux 2 (Ubuntu 22.04.5) distribution.

\subsection{Experiment setups}
\label{sec:setup}

\paragraph{Datasets.}
We validate our relighting quality by comparing it to previous methods on real and synthetic datasets from NRHints~\cite{zeng2023relighting} and RNA \cite{mullia2024rna}. We run all results on down-sampled datasets with $512\times512$ resolution and a maximum of 1000 training views. For synthetic data, all backgrounds are colored in black. 

\paragraph{Comparing methods.}
We select four representative NeRF-/GS-based relighting methods for comparison. For validating the relighting quality under point lights, we compare with NRHints~\cite{zeng2023relighting} and \rgs~\cite{bi2024rgs}. Furthermore, we compare with GS-IR~\cite{liang_2024_gsir} and Relightable 3D Gaussian~\cite{gao_2023_relightablegs} to validate the relighting under environment lights.

\paragraph{Metrics.}
We provide the peak signal-to-noise ratio (PSNR), SSIM, and perceptual similarity (LPIPS)~\cite{zhang_2018_lpips} values for comparison, to compare both pixel-wise error and visual differences. 

\subsection{Quality validation}
\label{sec:validation}

In Fig.~\ref{fig:vs_nrhints_real+syn}, we compare our relighting results with NRHints and \rgs under point lighting on real-world objects and synthetic scenes.  We provide renderings under novel views/lights and their difference maps for comparison.
Both \rgs and our method have finer details, especially for furry objects. Both NRHints and our method handle the shadow effects well, producing more solid and sharp shadow regions. We typically have lowest LPIPS and highest/second-highest PSNR and SSIM values, indicating the overall quality of our method. In terms of training time, both \rgs and our method are significantly faster (more than $20\times$) than NRHints.

In Table~\ref{tab:metrics}, we also report the statistics of NRHints, \rgs, and our method on a series of datasets. Our method has overall lower LPIPS and close SSIM values to \rgs, as we have better details. Since we also introduce the shadow cues and hybrid optimization, the shadow quality also increases the realism of our renderings. We are also competitive with NRHints in terms of pixel-wise errors with much lower training/rendering time cost, thanks to the efficiency of 3DGS and the flexibility of neural appearance models.

In Fig.~\ref{fig:relight_envmap}, we provide the relighting results under novel environment lighting and compare them with prior GS-based methods \cite{liang_2024_gsir, gao_2023_relightablegs}, and report the quantitative results across different datasets in Table \ref{tab:relight_envmap}. We run their methods with datasets under unknown environment lighting, and run our method with the same amount of training views under point lights, to be as fair as possible. After training all models, we relight them under the same novel environment map for quality comparison.
Our method produces closer appearance to the reference and more plausible shadow effects, achieving higher PSNR/SSIM and lower LPIPS values. Our method benefits from two aspects: the point-lit input images and the neural appearance model. Capturing with point lights helps us better decompose the lighting and materials, and the neural appearance model can handle complex light transport such as sub-surface scattering and hair fiber scattering, leading to overall better results.

We also provide additional validation and visualization of our learned geometry and shadows, the relighting quality, and the power of neural appearance in our supplementary. Please refer to them for more details.

\subsection{Ablation study}
\label{sec:albation}

\paragraph{Ablation of all components.}
In Fig.~\ref{fig:ablation-all}, we show the ablation of our model by gradually removing the depth refinement MLP, the shadow cue, and the deferred shading. The quality gap indicates the effectiveness and necessity of all the components of \method. 
We also provide the ablation study on network sizes in our supplementary.

\paragraph{Effect of depth refinement network.}
In Fig.~\ref{fig:ablation-d_mlp}, we show the significance of the depth refinement network. We compare the shapes and positions of the cast shadows via shadow mapping with/without the depth refinement network. As a result, there is obvious mismatching in positions with the ground truth if we remove the depth refinement network. The model can more accurately locate the shading points and generate more reasonable shadow cues with depth refinement. 
\subsection{Discussion and limitations}

\paragraph{Precision of our representation.}
\method shows less accuracy in terms of PSNR values in some scenes than NRHints. The main reason is that NRHints uses an SDF as a powerful prior, which gives very accurate shadows for objects that are clearly surface-like. Furthermore, NRHints uses larger networks than ours; we trade off between quality and computational overhead.

\paragraph{Geometry quality.}
Although we deploy the shadow cue to help the network predict better shadow appearances and present the depth refinement network to compensate for this inaccuracy, the shadow quality is still limited by the geometry reconstruction precision, since a perfect geometry reconstruction for soft-boundary objects is not trivial. Therefore, we suffer from this disadvantage like most GS-based approaches. 

\paragraph{Complex material effects.}
\method can handle objects with both hard surfaces and ill-defined shapes. However, since we use rasterization instead of ray marching, it is difficult for our model to handle highly reflective appearances. We also show a failure case in Fig.~\ref{fig:limitation}. The reflection of the checkerboard is blurry on the dice, and the reflected shadow is incomplete.

\section{Conclusion}
\label{sec:discussion}

In this paper, we have proposed \method for relighting both surface-based and soft-boundary objects under the 3DGS framework. The proposed neural Gaussian framework avoids assumptions on shading models, and the shadow cue helps produce sharp shadows, together with our hybrid optimization strategy. \method can render high-fidelity details and high-quality shadow effects, achieving real-time rendering with a significantly improved training ($1.3$ hours) and rendering ($60$ fps) performance, compared to prior work.

There are still many potential future research directions. For example, introducing reflection and refraction into the existing framework may be promising. Loosening the requirement on the lighting conditions and supporting more flexible capture setups would be valuable but also challenging. Another potential direction is to explore a more accurate definition of depths in a Gaussian splatting framework, further improving the shadow quality.

\section*{Acknowledgements}
We thank the reviewers for their valuable comments. This work is supported by the National Science Fund of China under Grant Nos. U24A20330, 62361166670, and 62172220.

{
    \small
    \bibliographystyle{ieeenat_fullname}
    \bibliography{main}
}

\clearpage
\setcounter{page}{1}
\maketitlesupplementary

We propose \method, a novel relightable asset with neural Gaussians. Without assumptions in the shading model and geometry types, we enable the relighting for both fluffy objects and surface-like scenes. 
In this supplementary material, we provide extra quality validation and metrics in \cref{sec:supp_validation}, and discuss the choice in hybrid optimization and the network size in \cref{sec:supp_discussion}.

\section{Additional validation}
\label{sec:supp_validation}

In Fig.~\ref{fig:vis_depth_shadow}, we visualize the obtained depth maps and shadow cues of our model of both real and synthetic objects. The depth map and shadow cues match well with our renderings and the ground truth. Our forward-deferred optimization strategy provides us with high-quality geometries, and together with reasonable shadow information, our model predicts close results to the reference.

In Table~\ref{tab:vs_3dgs}, we compare our neural appearance model to the vanilla 3DGS with SHs. We run both methods on datasets rendered with environment lighting and compare the NVS quality. Since the shadow cue and depth refinement MLP are disabled in our method, we only run forward shading for our method. Overall, our neural radiance representation provides more capacity and power in various scenes.

In Fig.~\ref{fig:rebuttal}, we demonstrate the effectiveness of our shadow cue by showcasing an example where the shadow quality dominates. We observe a significant quality improvement in the shadow by adding shadow cues into our model.

In Fig.~\ref{fig:move_light}, we move the light source towards and away from the object, showing the different lighting effects. Thanks to the shadow cue, our model shows robustness under different light conditions and can produce reasonable light effects.

In Fig.~\ref{fig:relight}, we show the relighting results of \method under moving point lights. Each column in the figure shows a different light direction. Our model can render scenes under novel lights with realistic appearance and high-quality details, and can properly model the self-shadowing effects. We suggest the reader refer to the supplementary video for more validation.

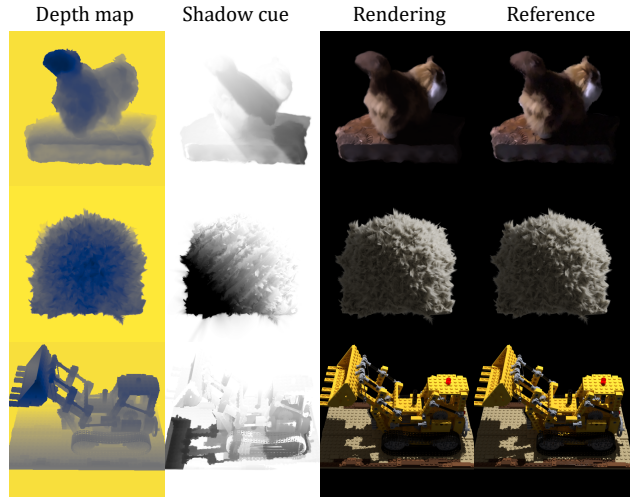
\myfigure{vis_depth_shadow}{vis_depth_shadow.pdf}{The visualization of depth maps and shadow cue maps of our model for different objects. The two-stage hybrid optimization strategy provides clear and accurate geometries, and the shadow cue also correctly reflects the visibility information.}

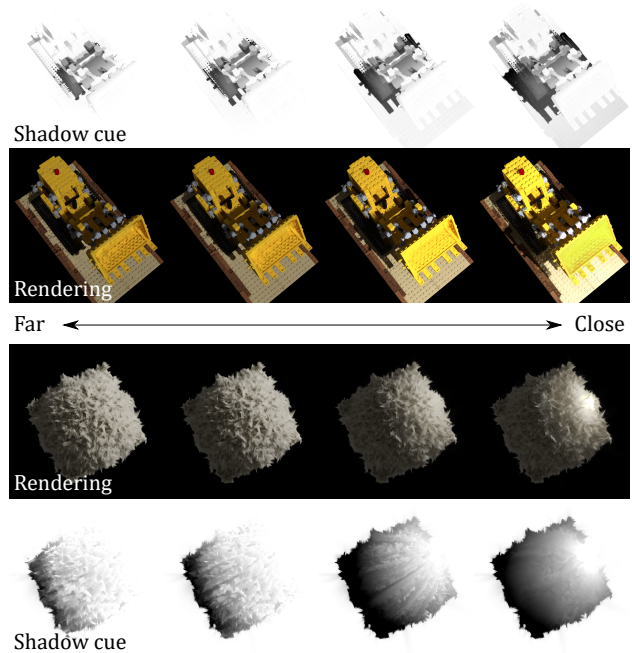
\myfigure{move_light}{move_light.pdf}{The renderings and visualization of shadow cues when moving the point light towards and away in the scene.  The shadow cues correctly reflect the movement of the light source, and our model produces plausible renderings.}

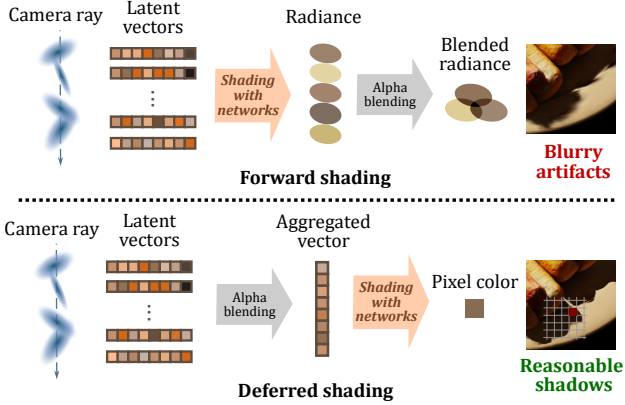
\myfigure{defer_shadow}{defer_shadow.pdf}{The difference in producing shadows between forward shading and deferred shading. For comparison, alpha blending usually leads to wrong and blurry shadows, while deferred shading provides more capability and flexibility to produce plausible colors for the blended image space features.}

\myfigure{ablation-defer}{ablation-defer.pdf}{The depth/shadow cue visualization and rendering comparison between forward and deferred shading. Forward shading produces better geometry, while deferred shading produces better shadows.}

\myfigure{rebuttal}{rebuttal.pdf}{
The quality difference at shadow regions with/without the shadow cues. With the shadow cues applied, the network can further improve the quality of dark pixels, as well as provide a clearer boundary for the shadow region.
}

\section{Additional discussion}
\label{sec:supp_discussion}

\paragraph{The choice in hybrid optimization.}
To improve the shadow quality and avoid blurry artifacts, we suggest a deferred shading process, regularizing the appearance of shadows in the image space. As shown in Fig.~\ref{fig:defer_shadow}, the appearance of shadows is not obtained by blending Gaussians, but directly decided in the image space, avoiding the artifacts brought by the blending operation. However, according to our observation, forward shading produces better geometry, while deferred shading leads to outliers and floaters. We show such observations in Fig.~\ref{fig:ablation-defer}. Therefore, we suggest a two-stage hybrid optimization strategy in the end, preserving both the geometry and shadow qualities.

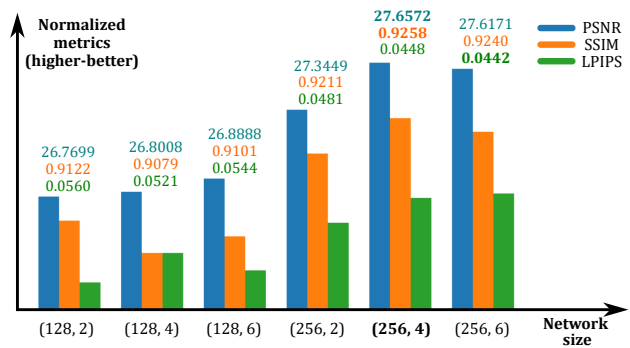
\myfigure{network_size}{ablation-network_size.pdf}{The comparison of variants with different network sizes. We test on \textsc{Furball} dataset. All metrics are normalized to be higher-better, and the best values of each metric are marked as \textbf{bold}. Our chosen configuration $(256, 4)$ achieves the balance between quality and complexity.}

\paragraph{Network sizes.}
In Fig.~\ref{fig:network_size}, we show the prediction accuracy with varying sizes of the neural Gaussian decoder. We use \textsc{Furball} for example, since this scene includes both complex appearance and shadows. All metrics are normalized so that higher values indicate better performance for ease of comparison. The variants are tagged by the number of hidden units and hidden layers, and our choice is $(256, 4)$. Our choice yields the best results among all variants, balancing between quality and computational complexity.

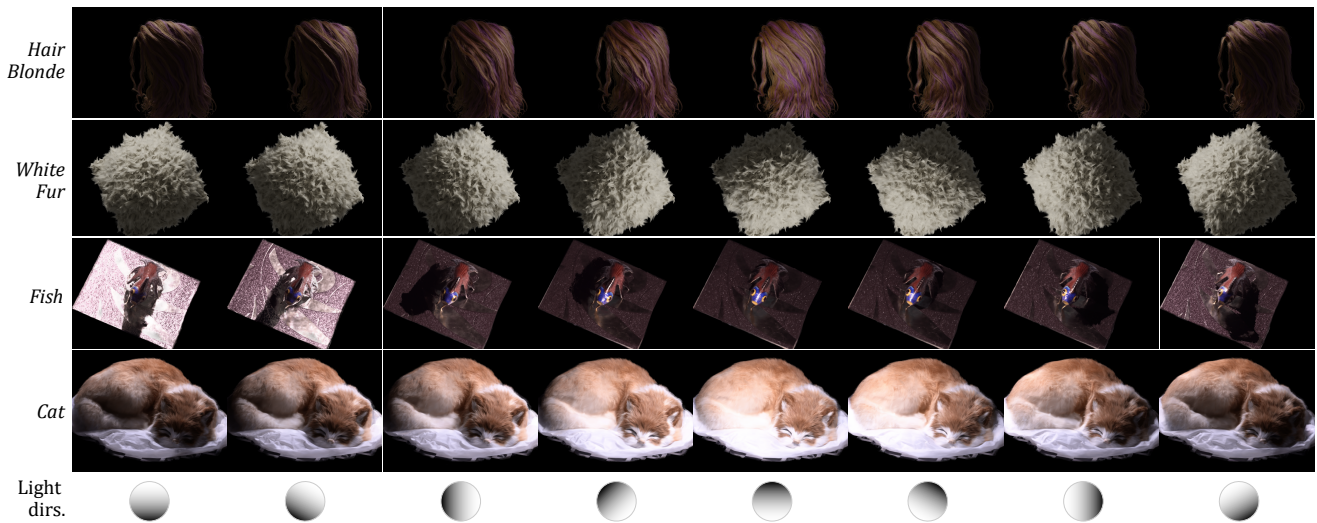
\mycfigure{relight}{relight.pdf}{The relighting results of \method. Each column shows a different point light direction. Our model can render scenes in novel views and lights with realistic appearance with high-quality details, and can properly model the self-shadowing effects.}

\begin{table*}
    \centering
    \caption{NVS Comparison of our neural appearance model and the vanilla SH-based 3DGS under static environment lighting. We provide (from left to right) PSNR$(\uparrow)$, SSIM $(\uparrow)$ and LPIPS $(\downarrow)$  for comparison, and the prevailing results are marked as \textbf{bold}. Our neural radiance representation is more flexible and powerful than the SHs in terms of NVS quality. Note that in this case, the shadow cue and depth refinement MLP are disabled in our method.}
    \label{tab:vs_3dgs}
    \begin{tabular}{p{0.10\textwidth} >{\centering\arraybackslash}p{0.25\textwidth} >{\centering\arraybackslash}p{0.25\textwidth}}
    \toprule
        Scene& Vanilla 3DGS& Ours (forward shading only)\\ 
    \midrule
 Armadillo & 45.8581 $\vert$ 0.9959 $\vert$ 0.0023 & \textbf{49.2875} $\vert$ \textbf{0.9976} $\vert$ \textbf{0.0009}\\
 CupPlane & 43.1947 $\vert$ 0.9957 $\vert$ 0.0021 & \textbf{47.3267} $\vert$ \textbf{0.9974} $\vert$ \textbf{0.0010}\\
 Ficus & 36.9734 $\vert$ 0.9937 $\vert$ 0.0038 & \textbf{39.892}1 $\vert$ \textbf{0.9964} $\vert$ \textbf{0.0019}\\
 Flowers & 36.0157 $\vert$ 0.9918 $\vert$ 0.0049 & \textbf{37.2329} $\vert$ \textbf{0.9941} $\vert$ \textbf{0.0036}\\
 HairBlue & 38.5114 $\vert$ 0.9766 $\vert$ 0.0197 & \textbf{39.5742} $\vert$ \textbf{0.9811} $\vert$ \textbf{0.0146}\\
 Hotdog & 35.3799 $\vert$ 0.9941 $\vert$ 0.0047 & \textbf{42.6534} $\vert$ \textbf{0.9972} $\vert$ \textbf{0.0015}\\ 
 Lego & 42.0764 $\vert$ 0.9962 $\vert$ 0.0021 & \textbf{44.9111} $\vert$ \textbf{0.9981} $\vert$ \textbf{0.0009}\\
 \midrule
 \textit{Average} & 39.7157 $\vert$ 0.9920 $\vert$ 0.0057 & \textbf{42.9826} $\vert$ \textbf{0.9946} $\vert$ \textbf{0.0035}\\
    \bottomrule
 \end{tabular}
\end{table*}

\end{document}